\documentclass[conference]{IEEEtran}
\IEEEoverridecommandlockouts
\usepackage{cite}
\usepackage{amsmath,amssymb,amsfonts}
\usepackage{algorithmic}
\usepackage{graphicx}
\usepackage{siunitx}
\usepackage{makecell}
\usepackage{textcomp}
\usepackage{xcolor}
\def\BibTeX{{\rm B\kern-.05em{\sc i\kern-.025em b}\kern-.08em
    T\kern-.1667em\lower.7ex\hbox{E}\kern-.125emX}}
\begin{document}

\title{UAV-CROWD: Violent and non-violent crowd activity simulator from the perspective of UAV}

\author{
    \IEEEauthorblockN{Mahieyin Rahmun\IEEEauthorrefmark{1}, Tonmoay Deb\IEEEauthorrefmark{1}, Shahriar Ali Bijoy\IEEEauthorrefmark{1}, Mayamin Hamid Raha\IEEEauthorrefmark{1}}
    \IEEEauthorblockA{\IEEEauthorrefmark{1}\textit{Department of Electrical and Computer Engineering, North South University, Dhaka, Bangladesh}
    \\\{mahieyin.rahman, tonmoay.deb, shahriar.ali, mayamin.raha\}@northsouth.edu}
}

\maketitle
\thispagestyle{plain}
\pagestyle{plain}

\begin{abstract}
Unmanned Aerial Vehicle (UAV) has gained significant traction in the recent years, particularly the context of surveillance. However, video datasets that capture violent and non-violent human activity from aerial point-of-view is scarce. To address this issue, we propose a novel, baseline simulator which is capable of generating sequences of photo-realistic synthetic images of crowds engaging in various activities that can be categorized as violent or non-violent. The crowd groups are annotated with bounding boxes that are automatically computed using semantic segmentation. Our simulator is capable of generating large, randomized urban environments and is able to maintain an average of 25 frames per second on a mid-range computer with 150 concurrent crowd agents interacting with each other. We also show that when synthetic data from the proposed simulator is augmented with real world data, binary video classification accuracy is improved by 5\% on average across two different models.
\end{abstract}


\begin{figure*}
\begin{minipage}[b]{0.24\linewidth}
  \centering
    \centerline{\includegraphics[width=\textwidth, height=4cm]{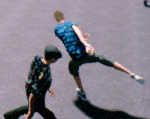}}
  \centerline{(a) Kicking}\medskip
\end{minipage}
\begin{minipage}[b]{0.24\linewidth}
  \centering
    \centerline{\includegraphics[width=\textwidth, height=4cm]{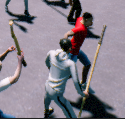}}
  \centerline{(b) Using weapons}\medskip
\end{minipage}
\begin{minipage}[b]{0.24\linewidth}
  \centering
    \centerline{\includegraphics[width=\textwidth, height=4cm]{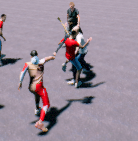}}
  \centerline{(v) Punching}\medskip
\end{minipage}
\begin{minipage}[b]{0.24\linewidth}
  \centering
    \centerline{\includegraphics[width=\textwidth, height=4cm]{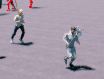}}
  \centerline{(d) Chasing}\medskip
\end{minipage}
\\
\begin{minipage}[b]{0.24\linewidth}
  \centering
    \centerline{\includegraphics[width=\textwidth, height=4cm]{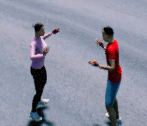}}
  \centerline{(e) Having conversation}\medskip
\end{minipage}
\begin{minipage}[b]{0.24\linewidth}
  \centering
    \centerline{\includegraphics[width=\textwidth, height=4cm]{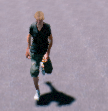}}
  \centerline{(f) Walking}\medskip
\end{minipage}
\begin{minipage}[b]{0.24\linewidth}
  \centering
    \centerline{\includegraphics[width=\textwidth, height=4cm]{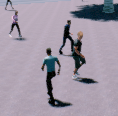}}
  \centerline{(g) Dispersion}\medskip
\end{minipage}
\begin{minipage}[b]{0.24\linewidth}
  \centering
    \centerline{\includegraphics[width=\textwidth, height=4cm]{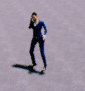}}
  \centerline{(h) Talking over phone}\medskip
\end{minipage}
  \caption{Simulated activities of crowd agents. \textbf{(a) - (d)} and \textbf{(e)-(h)} represent violent and non-violent activities, respectively}
  \label{fig:teaser}
\end{figure*}

\section{Introduction}
Crime rate has been rising in recent the years \cite{harp2020global}. With technological advancements, both law enforcement forces and criminals have become more effective. As such, automated surveillance techniques for constantly monitoring key locations is necessary. An important step in this automation is to recognize the type of action from video footage, which is an active topic of research \cite{deb2018machine}, \cite{zhou2018violence}. Human Activity Recognition (HAR) requires an extensive quantity of data. Furthermore, the usage of videos has gone up significantly in recent times. Added to it is the fact that the most common form of surveillance is done by surveillance cameras. The aforementioned factors combined pave the path towards revered video datasets focusing on action and attribute recognition, namely \cite{abu2016youtube,castanon2019out,monfort2019moments}. Additionally, surveillance cameras tend to provide a stationary point of view (PoV), resulting limited field of view (FoV).

Unmanned Aerial Vehicles (UAV) have become popular recently due to their precise and rapid movement capabilities. Although there are some datasets captured from aerial PoV such as UCF-ARG \cite{nagendran2010new} and Okutama-action \cite{barekatain2017okutama}, they focus on general human activity. In a recent work that focuses on violent human activity recognition \cite{singh2018eye}, the researchers use dataset that was collected by the themselves, which further proves the lack of such data. Besides, collecting real life data on violence can be threatening to the person trying to capture the activity. Hence, we would like to propose our solution to the aforementioned problems in the form of a simulator which is capable of:
\begin{itemize}
	\item Generating photo-realistic RGB images captured from the perspective of an UAV,
	\item Using procedural method for generating randomized urban environment,
	\item Generating various forms of ground truth images,
	\item Generating bounding box around crowd groups using semantic segmentation,
	\item Providing control over the velocity and altitude of the UAV via input e.g. joystick or RC.
\end{itemize}

The proposed crowd simulator is developed using Unreal Engine\footnote{Unreal Engine: https://unrealengine.com}, which is an open source game development engine. In the upcoming sections, we describe the relevant literature to our work, our methodology and some qualitative and quantitative results to demonstrate the effectiveness of the proposed simulator.

\section{Related Work}
The aim of our proposed simulator is two-fold. Firstly, we want to be able to simulate violent and non-violent crowd activity in a randomized manner so that we can obtain data related to it. And secondly, we want to be able to obtain the data from the perspective of an UAV, which will enable us to train models that can be applied in violence detection using UAVs as a medium of surveillance in key locations.

\subsection{Existing crowd simulation techniques} There has been a lot of related work in the field of crowd simulation, but most of them only involve neutral pedestrians \cite{kaup2006crowd,loscos2003intuitive}. \cite{curtis2016menge} deals with modeling how crowd should move in response to various triggers and changes in their surroundings by making use of Behavioral Finite State Machines. The focus of the authors mainly lies in developing realistic crowd movement by making use of goal selection, trajectory estimation and obstacle avoidance. These are built into Unreal Engine by default as part of its core library, as a result of which an Non-player character (NPC) can reach locations within the world from a starting location in a realistic way when controlled by the AI. \cite{aroor2017mengeros} is a framework for crowd simulation interconnecting robots, yet it is limited ground robots and is based on Menge framework for the crowd simulation process.
In \cite{ros2016synthia}, the authors develop a simulation environment using Unity, which is also a video game development. They aim to generate semantic segmentation images from real life urban scenarios captured in the form of RGB images. They show that the synthetic dataset improves the classification accuracy of the Deep Convolutional Neural Network (DCNN) when trained to generate semantic segmentation images.
\cite{wang2019learning} publish a dataset that is generated using the video game Grand Theft Auto V published by Rockstar Games\footnote{Rockstar Games: https://www.rockstargames.com/}. Rockstar Games allows third-party mods to be installed and run inside the game environment, which allowed the researchers to instantiate environment with crowd.

\begin{figure*}
\centerline{\includegraphics[width=\textwidth]{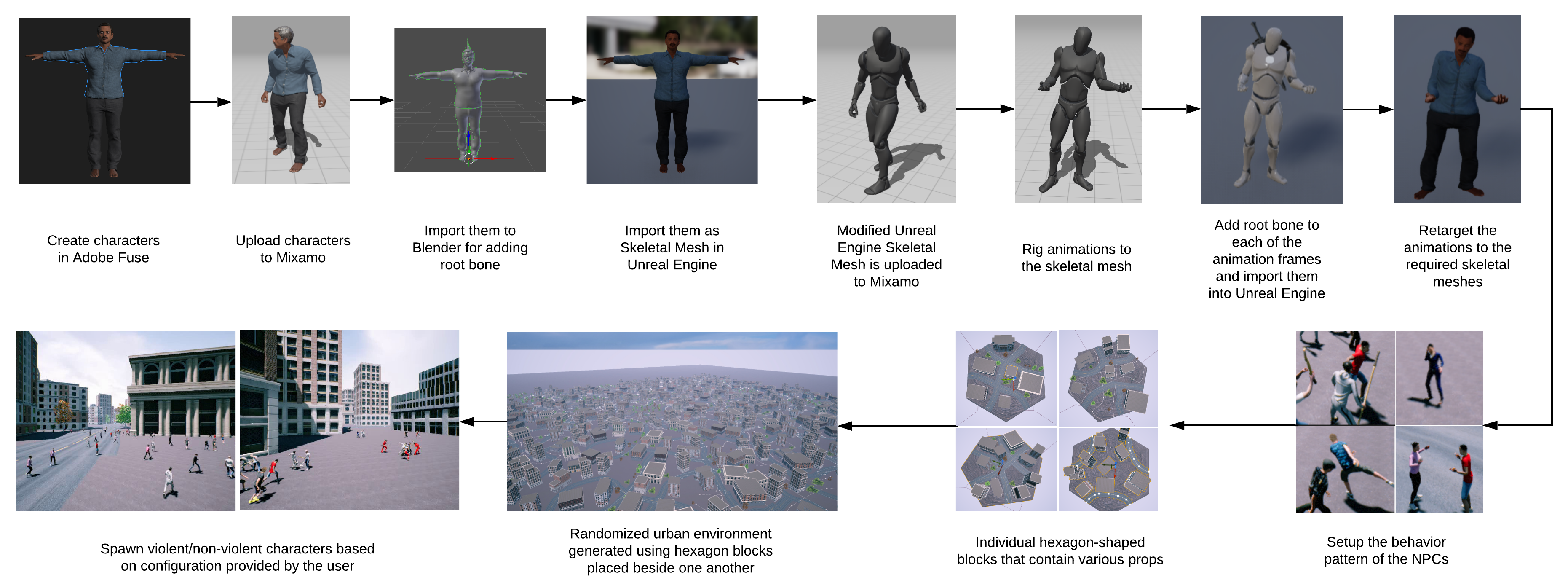}}
\caption{The pipeline for designing our proposed simulator}
\label{fig:proposedpipeline}
\end{figure*}

\subsection{Existing simulators and datasets} In existing datasets related to crowd, the PoV and quantity is limited. They mostly rely on RGB images only \cite{van2014nature}, \cite{olivares2015towards}. Moreover, in our particular case, the existence of videos from aerial viewpoint is scarce, and there aren’t many datasets that deal with violent behavior of crowd. \cite{mueller2016benchmark} propose a dataset which has 123 sequences captured from an UAV perspective and simulator built with Unreal Engine, but their focus is mostly single object tracking. \cite{kim2019synthesizing} propose a framework for generating synthetic data from Unreal Engine by making use of environment modeling, activity modeling and using the generated data for training machine learning models and data augmentation. But it is limited to a specific type of activity recognition which is pertaining to human interaction with cars. \cite{khirodkar2019domain} propose a simulator designed using Unreal Engine that is capable of generating real enough synthetic data of cars with a view to solving the domain shift problem \cite{pan2009survey}. \cite{muller2018sim4cv} propose a framework for generating environments with minimal user interaction and a pipeline for collecting image and ground truth data. \cite{tzimas2020leveraging} use Unreal Engine and character models from Mixamo\footnote{Mixamo: https://www.mixamo.com/} to train a deep reinforcement learning model to capture frontal faces of characters.

None of the above works focus specifically on modelling crowd behavior in violent and non-violent contexts in a procedurally generated simulation environment, which is the aim of our proposed simulator. We are inspired by \cite{tzimas2020leveraging} in terms of their level design technique, hence we extend it by introducing artificially intelligent characters and leverage procedural randomized level generation to meet the goals of our simulator.

\section{Simulator Design}
In this section, we discuss our approach in developing the simulator. The main workflow is in Unreal Engine using Airsim \cite{shah2018airsim} plugin to support UAV flight and image capture. We use AI Behavior Toolkit\footnote{AI Behavior Toolkit: https://www.unrealengine.com/marketplace/en-US/product/ai-behavior-toolkit} available in the Unreal Engine Marketplace\footnote{Unreal Engine Marketplace: https://www.unrealengine.com/marketplace/en-US/store} to model artificially intelligent Non-player Characters (NPC). For creating environments, we make use of various Unreal Engine Asset Packs available in the Unreal Engine Marketplace. The complete pipeline of the proposed system is shown in Figure \ref{fig:proposedpipeline}.

\subsection{Level Generation}
When generating the simulated world, we adopt a procedural level generationa approach which is an active research topic in the gaming community \cite{rogla2017procedural,maxim2018enhancing}. Our proposed framework is quite similar to existing grid based approaches using hexagon shaped blocks. The associated pseudocode for level generation and character spawning can be found in the \textbf{supplementary material}.

\subsection{Crowd Agents, Appearance and Behavior}
By crowd agents, we refer to the NPCs that are part of the simulation world and have the ability to navigate within the simulated world. We want the crowd agents to be fully dynamic and be able to interact intelligently with their surroundings. The interaction can vary depending on the context e.g. violent or non-violent. We accomplish such behavior modeling using AI Behavior Toolkit. To make the characters realistic we use Adobe Fuse\footnote{Adobe Fuse: https://www.adobe.com/products/fuse.html} and Mixamo for character animations.

\subsection{Data and Semantic Ground Truth Generation}
Once the simulator is launched, we use the Python API provided by Airsim to spawn an UAV in the simulated world, which can be controlled using API or joystick. We make API calls to request images from the render thread that captures the specified image types (RGB, Segmentation, Depth-Perspective etc.) and save them locally. The necessary calculations for generating segmentation and depth images are handled by the Airsim plugin. We include a demo video of the simulator tool in the \textbf{supplementary materials} for interested readers. We are also able to generate ground truth boxes around crowds by assigning a fixed color with RGB code (81, 13, 36) to the NPCs in the segmentation images, which are then processed to obtain ground truth for the corresponding RGB frames.

\begin{figure}[h]
\begin{center}
	\includegraphics[width=0.23\textwidth, height=3cm]{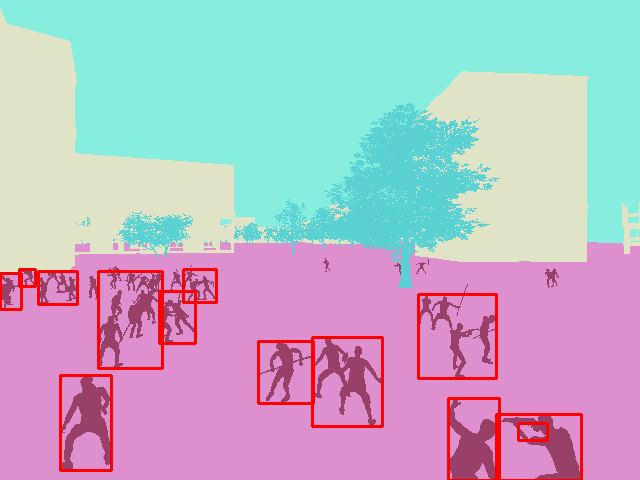}
	\includegraphics[width=0.23\textwidth, height=3cm]{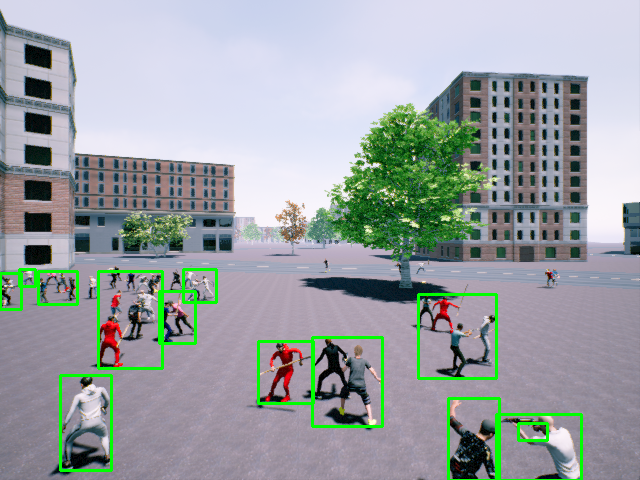}
	\includegraphics[width=0.23\textwidth, height=3cm]{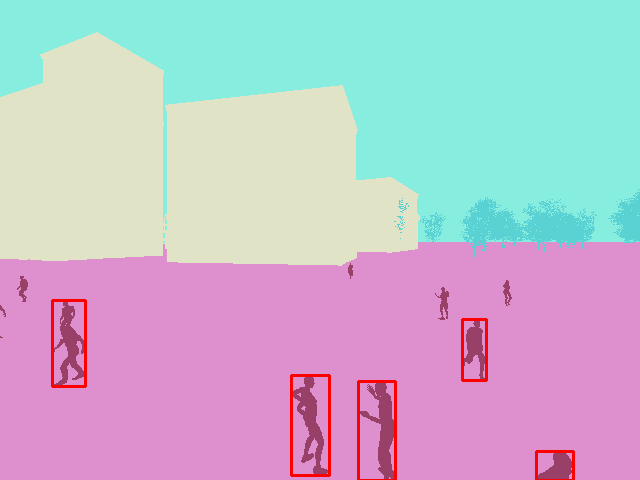}
	\includegraphics[width=0.23\textwidth, height=3cm]{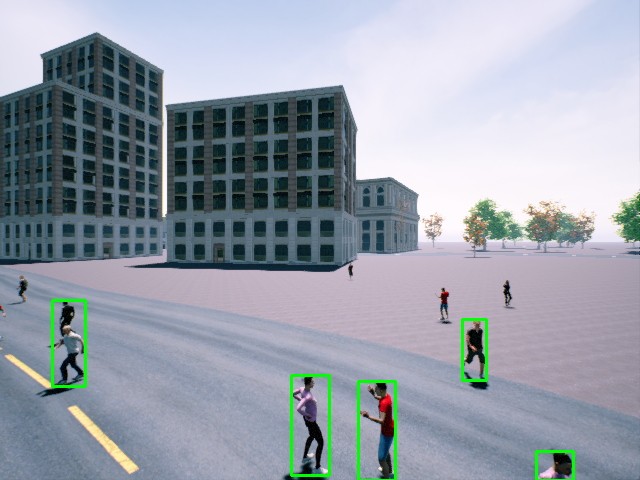}
\end{center}
   \caption{Ground truth is computed from the segmentation image (left column) and reapplied to the corresponding RGB frame (right column)}
\label{fig:gt}
\end{figure}

\begin{table*}
  \centering
  \caption{A comparison of the proposed simulator with other available crowd datasets and simulators}
  \label{table:comparison_with_other_datasets}
  \begin{tabular}{|c|c|c|c|c|c|c|c|}
    \hline
    \makecell{Name of Dataset} & \makecell{No. of\\Videos} & \makecell{No. of\\Frames} & Resolution & Is simulator? & \makecell{Has\\Segmentation\\Image?} & \makecell{Has\\Depth\\Image?} & \makecell{Contains\\Violence?}\\
    \hline
    \makecell{Crowd Collectiveness \cite{zhou2013measuring}} & 413 & 40,796 & 670x1000 & No & No & No & No \\
    \makecell{Data-driven crowd \cite{rodriguez2011data}} & 212 & 121,626 & 720x480 & No & No & No & No \\
    Violent Flows \cite{hassner2012violent} & 246 & 22,074 & 320x240 & No & No & No & Yes \\
    \makecell{WWW \cite{shao2015deeply}} & 10,000 & 8M  & 640x360 & No & No & No & Yes \\
    UCF \cite{ali2007lagrangian} & 46 & 18,196 & Variable & No & No & No & No \\
    Okutama-action \cite{barekatain2017okutama} & 43 & 77,365 & 3840x2160 & No & No & No & No \\
    \makecell{LCrowdV \cite{cheung2016lcrowdv}} & $>$1M  & $>$20M  & Any & Yes & No & No & No \\
    \textbf{Our Approach} & \textbf{Any} & \textbf{Any} & \textbf{Any} & \textbf{Yes} & \textbf{Yes} & \textbf{Yes} & \textbf{Yes}\\
    \hline
  \end{tabular}
\end{table*}

\section{Evaluation Metrics}
We evaluate the proposed simulator in a three-fold manner. First, we try to evaluate how many concurrent active NPCs can be rendered on screen without impacting the framerate too much. Next, we do visual comparison between some real world images and simulator-generated images. Finally, we perform quantitative comparison where we try to augment the simulator generated data with real world data and see if it improves model performance in binary video classification.

\subsection{Framerate}
\begin{figure}[t]
\begin{center}
    \includegraphics[width=\linewidth]{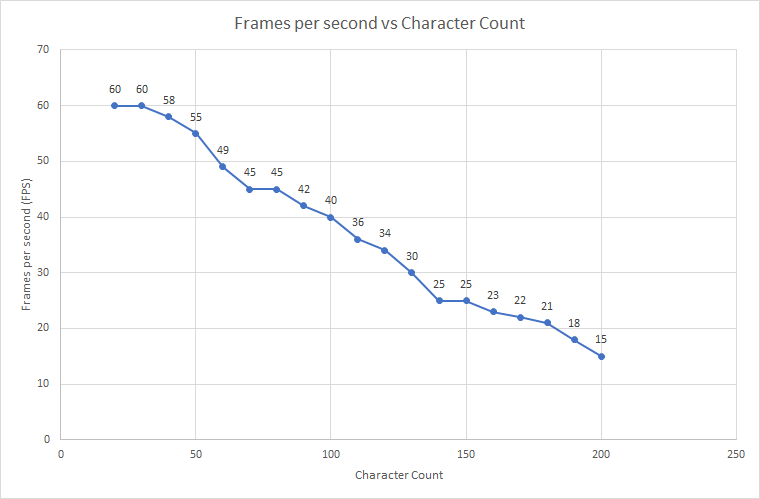}
\end{center}
  \caption{Average frame rate per second vs NPC count in simulation}
\label{fig:framesvscount}
\end{figure}

Our simulator is capable of maintaining 25 frames/second when rendering 150 characters simultaneously. The benchmark is done on a mid-range PC with an Intel Core i5-6500 @3.20 GHz, a RX 470 with 4GB VRAM and 16GB RAM. Figure \ref{fig:framesvscount} shows the corresponding graph that depicts the relationship between the number of characters and framerate.

\subsection{Qualitative Evaluation}
\begin{figure}[t]
\begin{minipage}[b]{0.32\linewidth}
  \centering
    \centerline{\includegraphics[width=\linewidth, height=2cm]{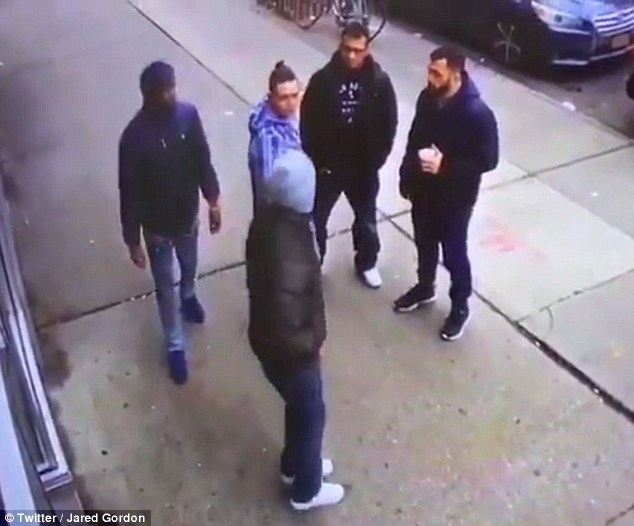}}
\end{minipage}
\hfill
\begin{minipage}[b]{0.32\linewidth}
  \centering
  \centerline{\includegraphics[width=\linewidth, height=2cm]{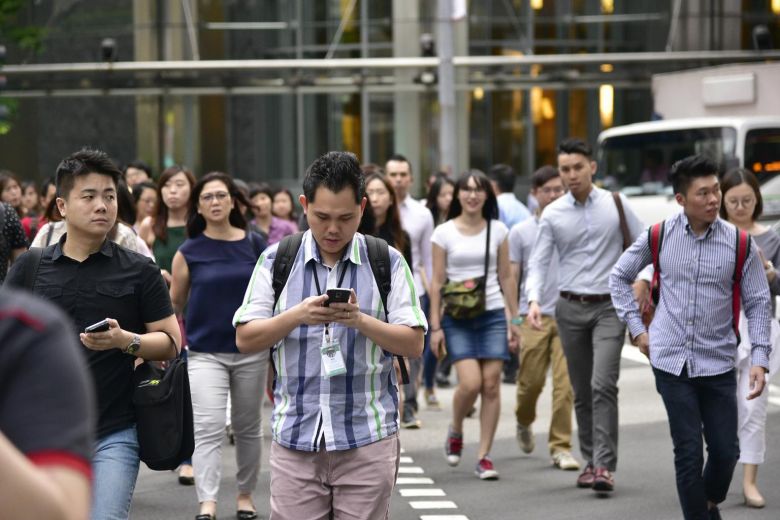}}
\end{minipage}
\hfill
\begin{minipage}[b]{0.32\linewidth}
  \centering
    \centerline{\includegraphics[width=\linewidth, height=2cm]{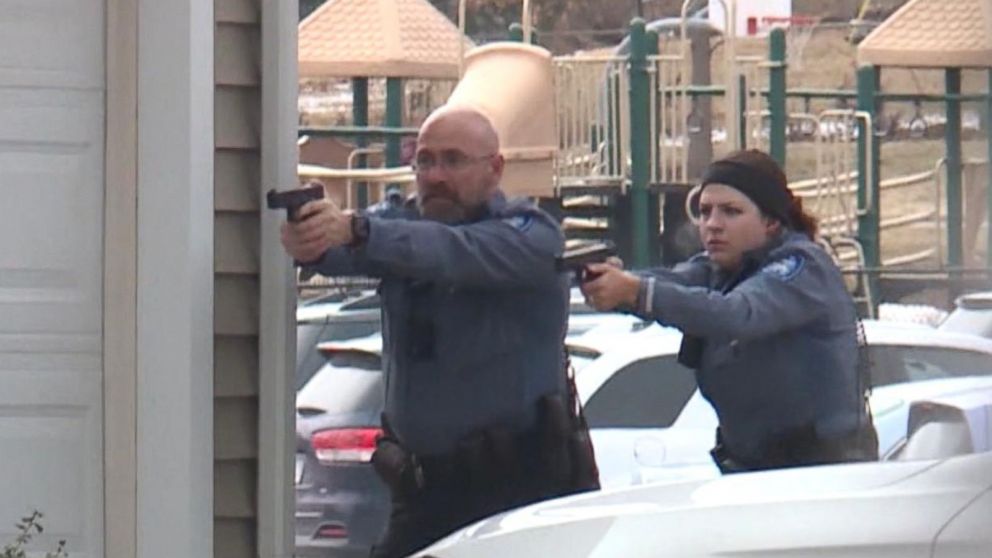}}
\end{minipage}
\hfill
\begin{minipage}[b]{0.32\linewidth}
  \centering
    \centerline{\includegraphics[width=\linewidth, height=2cm]{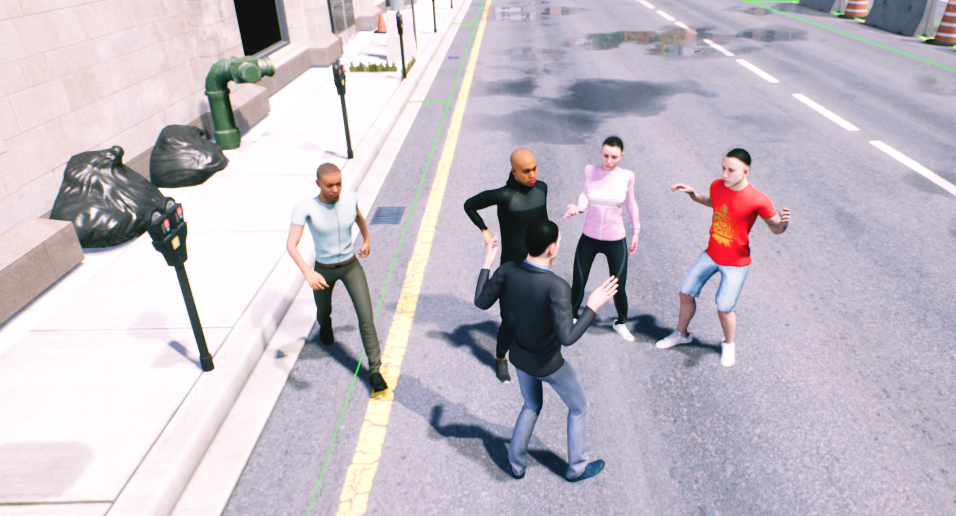}}
\end{minipage}
\hfill
\begin{minipage}[b]{0.32\linewidth}
  \centering
  \centerline{\includegraphics[width=\linewidth, height=2cm]{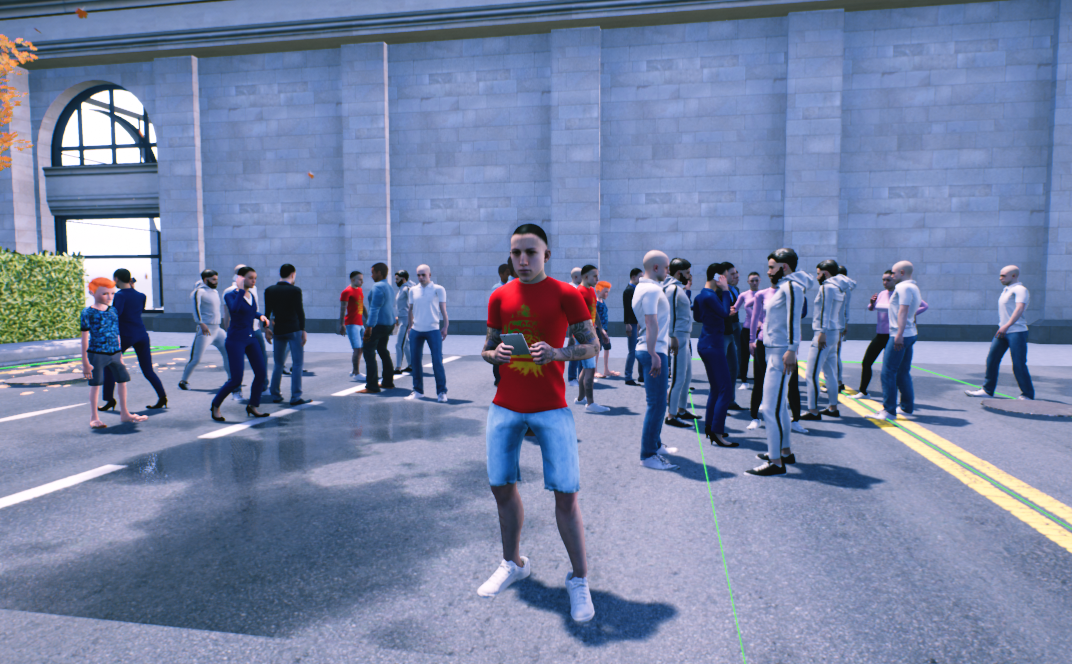}}
\end{minipage}
\hfill
\begin{minipage}[b]{0.32\linewidth}
  \centering
    \centerline{\includegraphics[width=\linewidth, height=2cm]{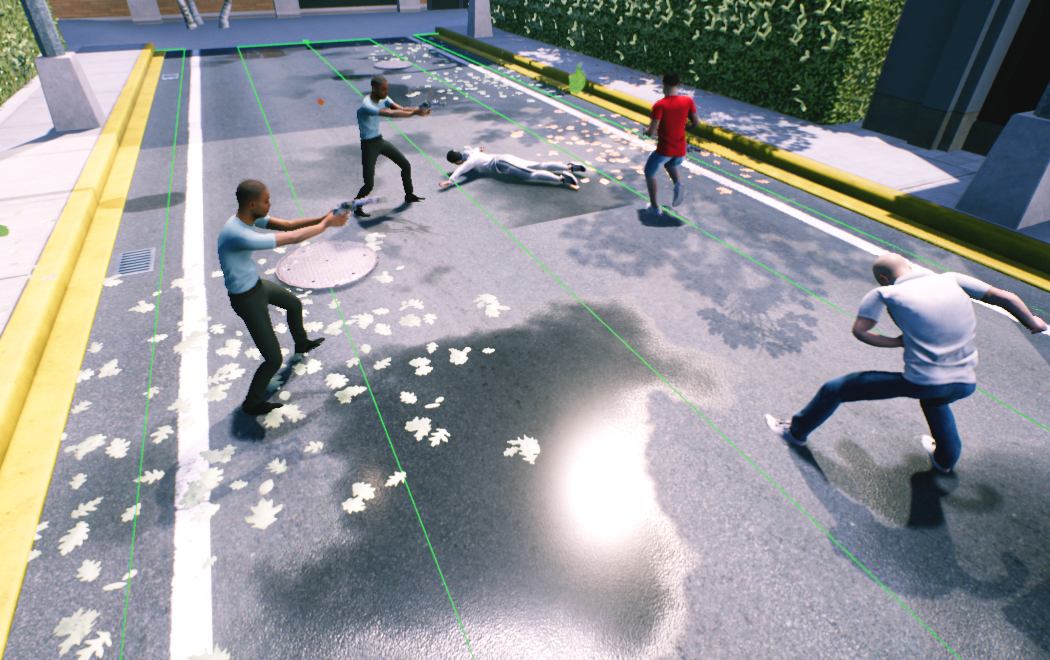}}
\end{minipage}
\caption{A visual comparison of real world images (top row) with synthetic data (bottom row).}
\label{fig:qualitativeanalysis}
\end{figure}

Figure \ref{fig:qualitativeanalysis} presents a qualitative comparison of the frames produced by the simulator with real world images. It is to be noted that the goal of the simulator is not to mimic every possible detail but to be able to simulate the crowd behavior as closely as possible. Character models with greater detail, more realistic and complex animations along with an increase in quantity of both will yield better and even more realistic synthetic data.

\subsection{Quantitative Evaluation}
In this section, we lay down the methodology in which we evaluate the data from the proposed simulator. We go over the datasets we use for our evaluation strategy and discuss the effects of augmenting the simulator data with the datasets in an endeavor to improve video classification performance.

\subsubsection{Dataset}
As we have mentioned earlier, even though there are various video datasets that have violent and non-violent sequences, crowd activity datasets are rare and most of those datasets are not captured from a UAV point-of-view. \cite{singh2018eye} introduce a dataset called Aerial Violent Individual (AVI), however, that dataset deals with images of violent individuals and not videos. \cite{barekatain2017okutama} propose a high resolution dataset that contains sequences captured from a UAV perspective, but it deals with human action recognition and contains no violent sequences. As such, we use the datasets mentioned below for evaluation. It is to be noted that most of the sequences in the real world video datasets are not captured from a UAV point-of-view.

\paragraph{Violent Flows \cite{hassner2012violent}}
The Violent Flows dataset is a classic dataset that has 246 total video sequences of various resolution and length, 123 of which are violent while the rest are non-violent.

\paragraph{Movie Fights \cite{nievas2011violence}}
This is yet another classic dataset that is composed of 200 video sequences equally divided into violent and non-violent categories collected from action movies.

\paragraph{Automatic Violence Detection in Videos \cite{Bianculli2020}}
This is a more recent dataset that has a total of 350 video clips. 120 among them are non-violent, while the rest are violent sequences.

\paragraph{Synthetic Dataset}
This is the dataset that is generated from our simulator. There are a total of 240 video sequences divided equally into violent and non-violent categories. All videos are captured from the perspective of an UAV at an approximately 3.5 meters above ground level in real world units within the Unreal Engine simulation. The violent sequences include punching, kicking, shooting and chasing, while non-violent sequences include talking, walking, dispersing and dancing. We augment this data with the real world datasets to find out whether adding it benefits classification performance.

\subsubsection{Preprocessing the datasets}
All videos, whether real world or synthetic are normalized to have a resolution of 640x480 at 30 frames per second, and are no longer than 10 seconds. Next, equal numbers of violent and non-violent sequences are chosen from the total dataset to ensure class balance. They are then divided into train, validation and test sets ensuring no data leakage among the sets. The test set is held out and not used in training or validation. For each dataset, 20\% of the data is kept for test set and another 20\% from the remaining data is kept as validation set. Table \ref{table:clipcount} shows the clip distribution across all datasets after preprocessing.

\begin{table}
    \centering
  \caption{Number of clips per dataset after preprocessing}
  \label{table:clipcount}
  \begin{tabular}{|c|c|c|c|}
    \hline
    \makecell{Name of\\Dataset} & Train & Validation & Test\\
    \hline
    Violent Flows & 158 & 38 & 50 \\
    Movie Fights & 128 & 32 & 40 \\
    \makecell{Automatic Violence Detection in Videos} & 154 & 38 & 48 \\
    Synthetic & 154 & 38 & 48 \\
    \hline
  \end{tabular}
\end{table}

\subsubsection{Network Architecture}
For training and inference, we use two models. All models are trained for 30 epochs on each set of data and then evaluated on the held out test data. All the models were trained using the MMAction2 toolbox \cite{mmaction2019}.

\paragraph{Temporal Segment Networks (TSN) \cite{wang2016temporal}} TSN is a state of the art model for action recognition which is based on the idea of formulating a temporal structure using temporal sampling and video-level supervision. We train the network on the datasets using a ResNet50 \cite{He2015} backbone. A learning rate of \SI{7.8e-5} is used.

\paragraph{Inflated 3D ConvNet (I3D) \cite{carreira2017quo}} I3D is another state of the art model that operates by expanding the dimension of filters and pooling layers of 2D ConvNet layers to 3D. Similar to TSN, we use a ResNet50 backbone and a learning rate of \SI{7.8e-5}.

\section{Experimental Results}
In this section we discuss the results obtained from our experiments. Moving forward, we refer to the datasets using the following acronyms: Violent Flows - \textbf{VF}, Movie Fights - \textbf{MF}, Automatic Violence Detection in Videos - \textbf{AVD} and Synthetic dataset - \textbf{S}.

\subsection{Individual datasets}
We use each of the datasets to establish a baseline and see how well the models perform when it is just trained on a single dataset and tested on test set from the same dataset. Results are shown in Table \ref{table:baseline}. It is seen that almost all models perform quite well on the datasets.

\begin{table}
  \centering
  \caption{Baseline performances of the models on the 4 datasets}
  \label{table:baseline}
  \begin{tabular}{|c|c|c|c|}
    \hline
    \makecell{Name of\\Dataset} & Model & Test Set & \makecell{Test\\Accuracy}\\
    \hline
    VF & \makecell{TSN\\I3D} & VF & \makecell{88.0\%\\83.8\%} \\
    \hline
    MF & \makecell{TSN\\I3D} & MF & \makecell{92.5\%\\88.0\%} \\
    \hline
    AVD & \makecell{TSN\\I3D} & AVD & \makecell{91.7\%\\87.5\%} \\
    \hline
    S & \makecell{TSN\\I3D} & S & \makecell{98.0\%\\100.0\%} \\
    \hline
  \end{tabular}
\end{table}

\subsection{Augmenting synthetic data with real world data}
Next, we start augmenting the synthetic data with real world data to see if it helps in improving classification performance. In Table \ref{table:data_augmentation}, we see that adding synthetic data nets us a few more predicitons correct across most datasets. We see a maximum gain of 8.2\% and a minimum gain of 2.1\%.

\begin{table}
  \centering
  \caption{Results after augmenting synthetic data with real world data}
  \label{table:data_augmentation}
  \begin{tabular}{|c|c|c|c|}
    \hline
    \makecell[l]{Name of\\Dataset} & Model & Test Set & \makecell{Test\\Accuracy}\\
    \hline
    VF + S & \makecell{TSN\\I3D} & VF & \makecell{94.0\%\\92.0\%} \\
    \hline
    MF + S & \makecell{TSN\\I3D} & MF & \makecell{95.0\%\\92.5\%} \\
    \hline
    AVD + S & \makecell{TSN\\I3D} & AVD & \makecell{93.8\%\\93.8\%} \\
    \hline
  \end{tabular}
\end{table}

\subsection{Augmenting real world data with real world data}
After that, we combine data from the datasets together to see how much improvement occurs if real world data is used instead of synthetic data. This is to evaluate if the models are performing well simply because of having more training data from the synthetic dataset. Table \ref{table:real_on_real} shows us that the performance is worse in some instances and see no improvements in others. This is because, the real world videos have random camera movements and jerky motion compared to the data from the simulator, which has more stable camera position. This contributes to the classifier failing to make predictions correctly, since the quality of the data is not as good.

\begin{table}
  \centering
  \caption{Results after increasing real world available data and using no synthetic data}
  \label{table:real_on_real}
  \begin{tabular}{|c|c|c|c|}
    \hline
    \makecell{Name of\\Dataset} & Model & Test Set & \makecell{Test\\Accuracy}\\
    \hline
    VF + MF & \makecell{TSN\\I3D} & VF & \makecell{88.0\%\\96.0\%} \\
    \hline
    VF + MF & \makecell{TSN\\I3D} & MF & \makecell{92.5\%\\87.0\%} \\
    \hline
    AVD + VF & \makecell{TSN\\I3D} & VF & \makecell{88.0\%\\85.0\%} \\
    \hline
    AVD + MF & \makecell{TSN\\I3D} & AVD & \makecell{93.8\%\\95.8\%} \\
    \hline
    AVD + VF & \makecell{TSN\\I3D} & MF & \makecell{88.0\%\\90.0\%} \\
    \hline
  \end{tabular}
\end{table}

\subsection{Testing domain difference}
Finally, we switch up the domains of the data. That is, we use models trained on one dataset on test data from another dataset. Table \ref{table:domain_shift} illustrates the results. This is where the domain difference of the simulated data and real world data is evident, as all models perform really poorly. We intend to address this issue by experimenting with domain adaptation techniques to see if the results can be improved in a future work.

\begin{table}
  \centering
  \caption{Results of domain shift}
  \label{table:domain_shift}
  \begin{tabular}{|c|c|c|c|}
    \hline
    \makecell{Name of\\Dataset} & Model & Test Set & \makecell{Test\\Accuracy}\\
    \hline
    VF & \makecell{TSN\\I3D} & S & \makecell{43.0\%\\37.8\%} \\
    \hline
    MF & \makecell{TSN\\I3D} & S & \makecell{65.0\%\\53.8\%} \\
    \hline
    AVD & \makecell{TSN\\I3D} & S &  \makecell{48.0\%\\45.5\%} \\
    \hline
    S & \makecell{TSN\\I3D} & VF & \makecell{50.0\%\\56.8\%} \\
    \hline
    S & \makecell{TSN\\I3D} & MF & \makecell{48.0\%\\52.0\%} \\
    \hline
    S & \makecell{TSN\\I3D} & AVD & \makecell{50.0\%\\50.0\%} \\
    \hline
  \end{tabular}
\end{table}

\begin{figure}[t]
\includegraphics[width=\linewidth]{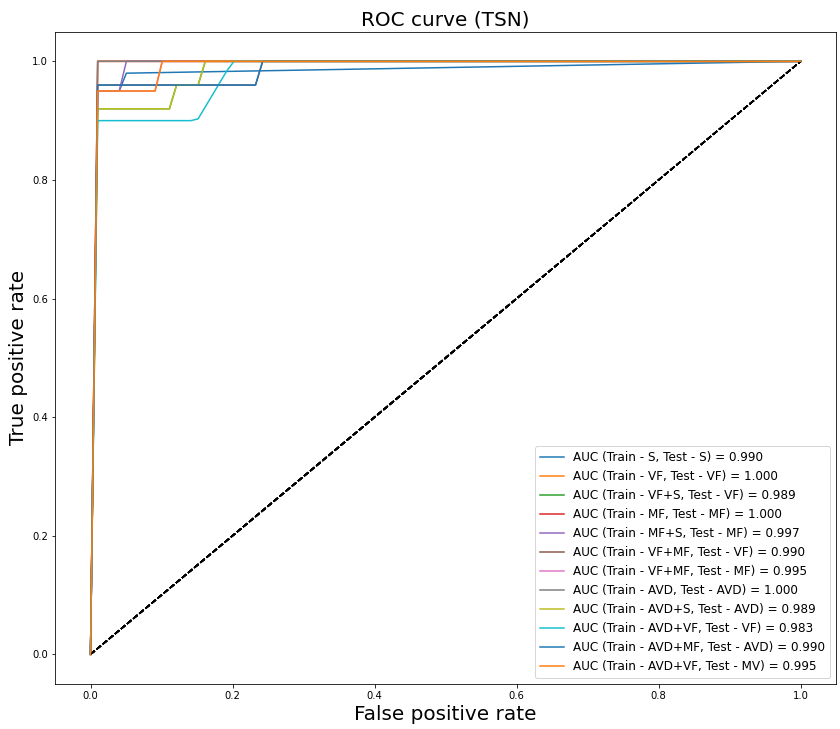}
\includegraphics[width=\linewidth]{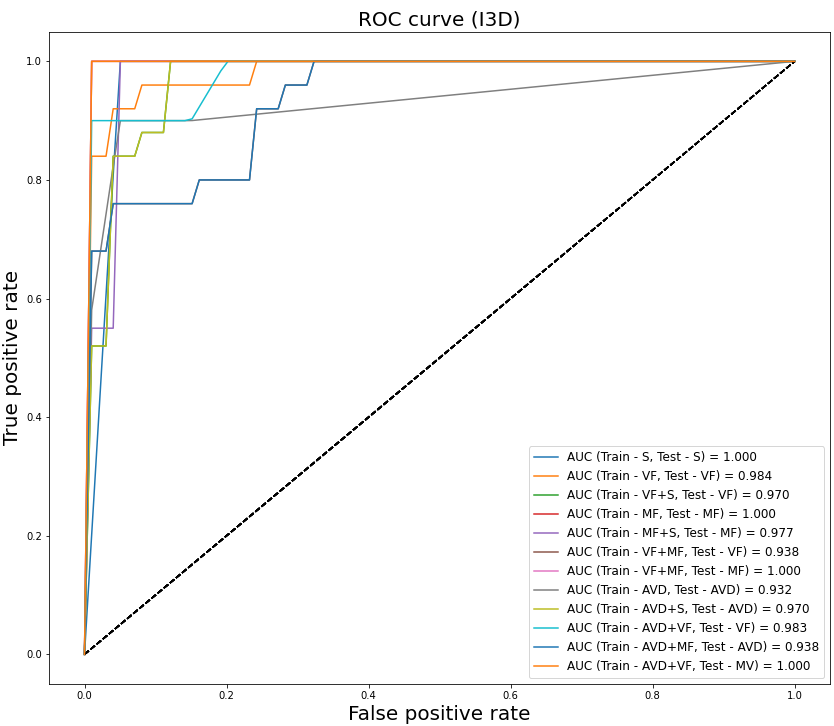}
\caption{ROC curves of the models when trained with various combinations of the mentioned datasets}
\label{fig:roc}
\end{figure}

Overall, we are able to confirm that the data from the simulator is benefiting the classification accuracy in most instances. The associated Receiver Operating Characteristics curves for TSN and I3D models are shown in Figure \ref{fig:roc}.

\section{Limitations, Conclusion and Future Work}
We propose a simulator that is capable of producing photo-realistic image sequences very near to real world environment in the context of violent and non-violent crowd activity captured from the perspective of an UAV. Leveraging a procedural level generation approach, we are able to create randomized urban environment, deploy AI-controlled NPCs in it and start generating data. Using segmentation images, we are able to provide bounding box annotations for the crowd in a particular frame automatically without the need of human intervention. Our work is presented as a baseline which is open for a wide range of extensions. Currently, the simulator lacks in the number of character models and diversified animations, which could be a point of argument that the data from the simulator is simplistic. If the number of character models and animations could be increased, it can greatly add to the realism of the simulated environment. Adding vehicles, destructible and flammable objects will open up room for more complex crowd behavior modelling. Additionally, we would like to extend our work in the future by introducing instance segmentation to the simulator, which will allow us to obtain bounding box data for individual NPCs, as well as generate videos from the perspective of a moving UAV. We also would like to experiment with domain adaptation techniques and see if the classification results can be further improved.

\bibliographystyle{IEEEtran}
\bibliography{references}

\end{document}